\documentclass[11pt,a4paper]{article}
\usepackage[hyperref]{acl2020}
\usepackage{times}
\usepackage{multirow}
\usepackage{diagbox}
\usepackage{amsmath}
\usepackage{latexsym}
\usepackage{graphicx}
\usepackage{amssymb}
\usepackage{booktabs}
\usepackage{color}
\usepackage{makecell}
\usepackage{subfigure}
\graphicspath{ {./images/} }

\usepackage{microtype}

\aclfinalcopy % Uncomment this line for the final submission
\title{Sequential Sentence Matching Network for \\
Multi-turn Response Selection in Retrieval-based Chatbots}

\author{Chao Xiong, Che Liu, Zijun Xu, Junfeng Jiang, Jieping Ye \\
  AI Labs, Didi Chuxing\\
  \texttt{ \{xiongchao,liuche,xuzijun,jiangjunfeng\_i,yejieping\}@didiglobal.com}
 }

\date{}

\begin{document}
\maketitle
\begin{abstract}
Recently, open domain multi-turn chatbots have attracted much interest from lots of researchers in both academia and industry. The dominant retrieval-based methods use context-response matching mechanisms for multi-turn response selection. Specifically, the state-of-the-art methods perform the context-response matching by word or segment similarity. However, these models lack a full exploitation of the sentence-level semantic information, and make simple mistakes that humans can easily avoid. In this work, we propose a matching network, called sequential sentence matching network (S2M), to use the sentence-level semantic information to address the problem. Firstly and most importantly, we find that by using the sentence-level semantic information, the network successfully addresses the problem and gets a significant improvement on matching, resulting in a state-of-the-art performance. Furthermore, we integrate the sentence matching we introduced here and the usual word similarity matching reported in the current literature, to match at different semantic levels. Experiments on three public data sets show that such integration further improves the model performance.
\end{abstract}

\section{Introduction}
Researchers have shown great interests in open domain multi-turn chatbots.
In general, there are two lines of approaches, the first one is retrieval-based~\citep{Tan:2015,Yan:2016,Zhang:2018b,Zhou:2018,Tao:2019a,Chen:2019,Wang:2019}, and the second one is generation-based~\citep{Serban:2016,Lowe:2017,Zhang:2018a}.
Retrieval-based approaches retrieve several candidates through searching a given database and then select the best response from these candidates using matching models.
Generation-based approaches use an encoder-decoder framework to generate responses word by word.
Generally, the former approaches are better in terms of the syntactic correctness, the diversity, and the length of the responses. The latter one tends to generate short and generic responses, which we call ``safe'' responses.
Therefore, retrieval-based approaches have been widely used in the industry such as the E-commerce assistant AliMe~\citep{Li:2018} serving on Taobao\footnote{https://www.taobao.com/} and the XiaoIce~\citep{Shum:2018} implemented by Microsoft \footnote{https://www.microsoft.com/}.
We focus on retrieval-based chatbots in this study.

The early retrieval-based models treat the context as a whole to match responses~\citep{Lowe:2015,Tan:2015,Yan:2016,Zhou:2016} while the recent ones use each utterance in the context to match the candidate response and then aggregate the matching results to choose a response~\citep{Wu:2017,Zhang:2018b,Zhou:2018,Tao:2019a,Tao:2019b}.
The recent ones work better because they retain more information of each turn, without compressing it into a single highly abstract vector.

Specifically, state-of-the-art methods follow a representation-matching-aggregation framework~\citep{Wu:2017,Zhang:2018b,Tao:2019a,Yuan:2019}.
In the matching stage they match each utterance with the candidate response by word-level or segment-level similarity matrix.
However, these matrices cannot fully reflect the sentence-level semantic information, and thus these methods make simple mistakes when similar words from the utterances appear in the negative candidates.
We show such a case in Table~\ref{table:case} to demonstrate this problem.

\begin{table}
    \tiny  %%\scriptsize %%\footnotesize %%\small %%\normalsize %%\large %%\Large %%\LARGE %%\huge %%\Huge
    \centering
    \begin{tabular}{c|l|c|c}
        \hline \textbf{Turns} & \textbf{Dialogue Context}  & \textbf{IOI} & \textbf{S2M} \\ 
        \hline
        Turn-1 & C: Please note that I've placed an order. \\
        Turn-2 & R: Please make sure your {\color{blue}shipping address} is correct. \\
        Turn-3 & \makecell[l]{C: Yeah it's correct. Btw, I have saved you as one of \\
        \quad my Saved Sellers. You said you would give free  \\
        \quad {\color{red}gift} for doing that. Do you still give now? } \\
        Turn-4 & R: Yes. \\
        Turn-5 & C: What {\color{red}gift} do you give? \\
        \hline
        Resp-1 & {\color{red}Wet tissues}. (True) & 0.86 & \textbf{0.98} \\
        \hline
        Resp-2 & \makecell[l]{Please make sure your {\color{blue}shipping address} is correct. 
        \\ (False)} & 0.56 & 0.01 \\
        \hline
        Resp-3 & \makecell[l]{There isn't any free {\color{red}gift}. Please make sure your \\
        {\color{blue}shipping address} is correct. (False)} & \textbf{0.98} & 0.91 \\
        \hline
    \end{tabular}
    \caption{\label{table:case} An case from E-commerce Dialogue Corpus. ``C" means customer and ``R" means customer service representative. IOI selects Resp-3 as the response because it has more matched words and phrases with the conversation context, such as ``gift", ``shipping address", while ``shipping address" is just a repetition, not a relevant phrase. S2M selects the correct response Resp-1, because sentence matching mechanism considers the semantics of the whole sentence then eliminates such kind of mistakes.}
\end{table}

To address this issue, we propose a sequential sentence matching network (S2M).
Instead of relying solely on word or segment similarity in matching, we calculate the similarity between a given utterance and response based on sentence-level matching.
It eliminates the kind of mistakes shown in Table \ref{table:case}.
Moreover, we design an effective mechanism to integrate sentence-level matching and word-level matching, so that we can take full advantage of different levels of semantic information and further improve the model performance.

We conduct experiments on three data sets: the Ubuntu Dialogue Corpus~\citep{Lowe:2015}, the Douban Conversation Corpus~\citep{Wu:2017}, and the E-commerce Dialogue Corpus~\citep{Zhang:2018b}.
The results show that the sentence matching mechanism reduces word matching errors and S2M significantly outperforms the state-of-the-art on the three corpora.
Experiment results also show that S2M performs better on longer context (for example, more than 10 turns) comparing to the models that match on word-level or segment-level.

Our contributions in this paper are four-folds:
(1) Proposal of a sequential sentence matching network in the context-response matching problem;
(2) Proposal of an effective integration method to incorporate the sentence matching mechanism and the word or segment matching mechanism;
(3) Empirical results show that our model significantly outperforms the state-of-the-art baselines on benchmark data sets;
(4) Empirical results also show that our model is more competent to handle long dialogues.

\begin{figure*}[ht]
    \centering
    \includegraphics[width=0.95\textwidth]{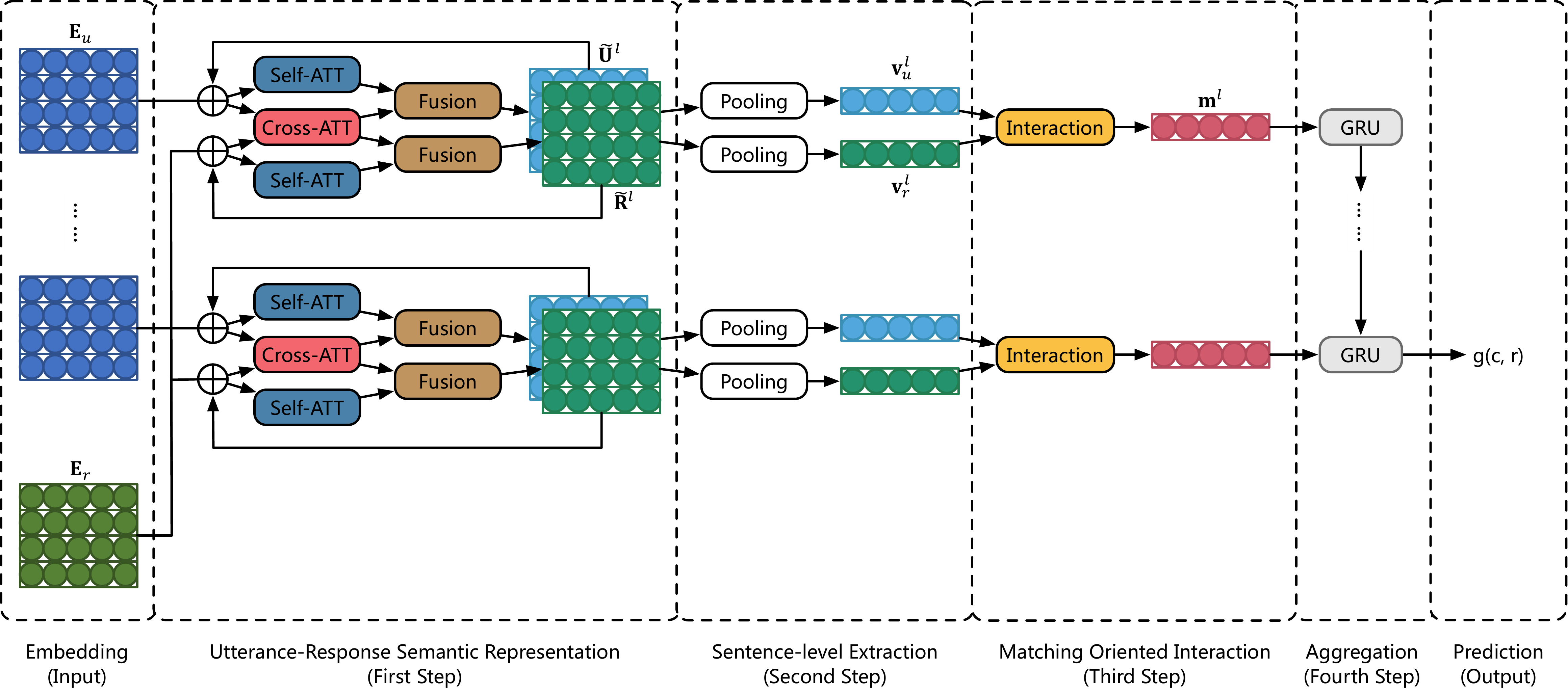}
    \caption{S2M model architecture.}
    \label{figure:model}
\end{figure*}

\section{Related work}
Open domain chatbot is one of the most important fields in NLP, specially on Dialogue System.
As we mentioned in the introduction, generation-based and retrieval-based approaches are two lines of approaches.
As for the retrieval-based, there are also two lines of approaches.
The first line of approaches encodes multiple utterances of the context into a single context vector and uses the vector to match responses.
~\citet{Lowe:2015} splice utterances into one sentence and then encode it with LSTM~\citep{Hochreiter:1997}.
~\citet{Zhou:2016} take each utterance as a unit and use hierarchical GRU to encode utterances into a context vector to catch utterance-level discourse information.
~\citet{Gu:2019} consider an interaction between the context and the response in order to generate more descriptive representation, and use an attentive hierarchical recurrent encoder to characterize the representation.

The second line of approaches matches a candidate response with every utterance of the context, and get a final match score by aggregation to decide whether it is a proper response. 
~\citet{Wu:2017} propose a representation-matching-aggregation framework, which firstly encodes each utterance and response, and then matches them by word-level and segment-level similarity matrix, and finally aggregates each turn's matching result by RNN.
The recent works follow this framework but make some modification.
~\citet{Tao:2019a} update the representation module by fusing multiple representations.
~\citet{Zhou:2018} and ~\citet{Tao:2019b} update both representation and matching modules.
~\citet{Zhou:2018} also consider the interaction between utterance and response and use stacked self-attention and cross-attention to perform the interaction.
~\citet{Tao:2019b} propose an interaction-over-interaction model which lets utterance-response interaction go deeper.
Other works show that utterances in the context are of different levels of importance for response matching.
~\citet{Zhang:2018b} hypothesize that the last utterance of the context is the most important.
It calculates the representations of the last utterance and of the other utterances respectively and concatenates them to obtain the context representation.
~\citet{Yuan:2019} propose a multi-hop selector network to select relevant utterances as the useful context and disregard the others.

However, the works mentioned above rely on word-level or segment-level similarity to perform matching, which actually does not capture the sentence-level semantic information. In this paper, we improve them by real sentence matching mechanism.

\section{Problem Formalization}
Suppose that we have a set of human dialogue sessions $\mathcal{D}=\{ (c_{k}, r_{k}, y_{k}) \}_{k=1}^{K}$, where $c_{k}=\{u_{1}, u_{2}...u_{t}\}$ is the context with $t$ turns of utterances and $r_{k}$ is the response, $y_{k}$ is the corresponding label with $y_{k}=1$ indicating positive sample and $y_{k}=0$ indicating negative sample.
Our goal is to learn a matching model $g(c, r)$ from $\mathcal{D}$, which can be used to evaluate the matching degree between any given $c$ and $r$ in practical use.

\section{Sequential Sentence Matching Network}
We propose a sequential sentence matching network (S2M) to model $g(c, r)$.
Figure \ref{figure:model} illustrates an overview of S2M.
S2M generally follows the representation-matching-aggregation framework, but introduces sentence matching mechanism in the matching process.
% 这里的matching process和sentence matching mechanism要统一一下是哪个模块, 关系到图和subsection. 其实也可以把2 3步概括认为是matching process.
Firstly, it pairs each utterance in the context with the response and transforms each pair into a sequence of representations through $L$ blocks with same structure.
Then for each block, the two new representations are pooled and interacted with each other to generate matching features.
Finally, all the matching features across turns are aggregated to obtain the matching score.

We further study the effect of integrating matching at word-level and sentence-level, and explore three different integration strategies.
We verify that matching at word-level and sentence-level are complementary, and integration of them significantly helps our model to select the best response.
We will describe the model in detail.

\subsection{Embedding Layer}
For a given utterance $u$ in the context and a given response $r$ as two sequences of words, the embedding layer converts $u$ and $r$ into the vector form ${\bf{E}}_{u}$ and ${\bf{E}}_{r}$ by looking up $\bf{P}$, where ${\bf{P}} \in \mathbb{R}^{d\times |V|}$ is the pre-trained Word2Vec \citep{mikolov2013distributed} on $\mathcal{D}$, $d$ is the dimension of the word embedding and $\left | V \right |$ is the size of the vocabulary.

\subsection{Semantic Representation Block}
\label{section42}
\begin{figure}[ht]
    \centering
    \includegraphics[width=0.45\textwidth]{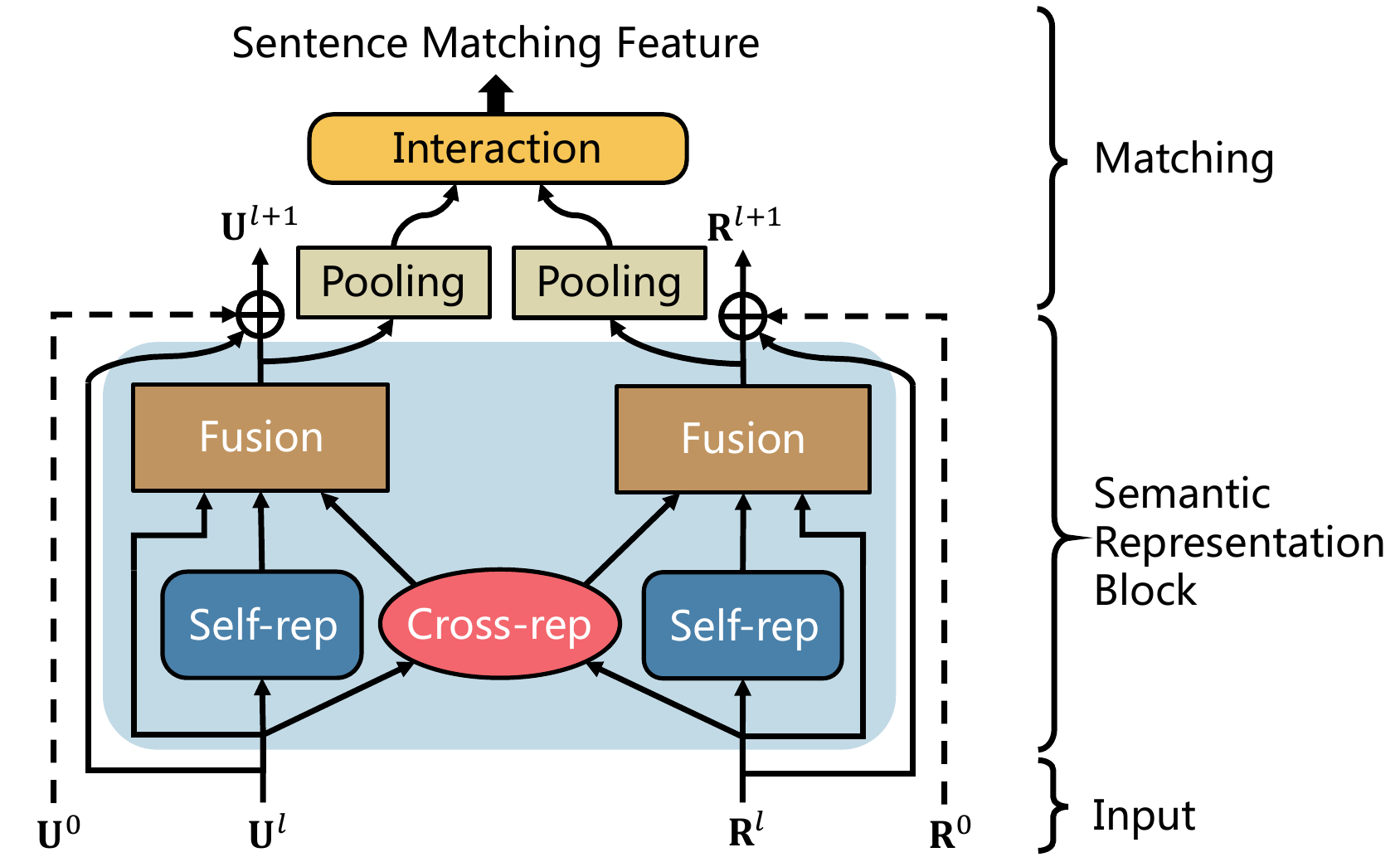}
    \caption{An overview of semantic representation block. First, we calculate the self-representations of the utterance (U) and the response (R) respectively and their cross-representations. Second, we fuse the representations to obtain two new representations. Finally, we match them by pooling and interaction.}
    \label{figure:sentence_block}
\end{figure}

A semantic representation block takes an utterance and a response as input, and generates two representations.
As shown in Figure \ref{figure:sentence_block}, such block consists of three components: self-representation module, cross-representation module and fusion module.
The self-representation module models the intra-sentence word relations for the utterance and the response respectively, and the cross-representation module models the cross-sentence relations between the utterance and the response.
The fusion module is then applied to fuse input embeddings, the self-representation results and the cross-representation results to form new representations.

\subsubsection{Self-Representation}
We employ 1D-CNN \citep{kim2014convolutional} as the implementation of the self-representation module.
Let ${\bf{U}}^{l}$ and ${\bf{R}}^{l}$ be the input of the $l$-th block where ${\bf{U}}^{0}={\bf{E}}_{u}$ and ${\bf{R}}^{0}={\bf{E}}_{r}$.
1D-CNN extracts contextual features from ${\bf{U}}^{l}$ and ${\bf{R}}^{l}$ by performing a one dimensional convolutional neural network with multiple filters, which is formulated as:
\begin{align}
  \bar{{\bf{U}}}^{l}_{i j} &= f ( {\bf{W}}_{i} \cdot {\bf{U}}^{l}_{j:j+h-1} ) + b_{i} \\
  \bar{{\bf{R}}}^{l}_{i j} &= f ( {\bf{W}}_{i} \cdot {\bf{R}}^{l}_{j:j+h-1} ) + b_{i}
\end{align}
where ${\bf{W}}_i\in \mathbb{R}^{h\times d}$ is the $i$-th filter with height $h$ and width $d$, $j$ is the beginning position of each possible window in ${\bf{U}}^{l}$ and ${\bf{R}}^{l}$, $f$ is a non-linear activation function and $b_{i}$ is a bias term. In our model, we use ReLU as the non-linear activation function.
All the contextual features of the $i$-th filter are then concatenated to form feature maps:
\begin{align}
  \bar{{\bf{U}}}^{l}_{i} = [\bar{{\bf{U}}}^{l}_{i1}, \bar{{\bf{U}}}^{l}_{i2}, \cdots,\bar{{\bf{U}}}^{l}_{i j}] \\
  \bar{{\bf{R}}}^{l}_{i} = [\bar{{\bf{R}}}^{l}_{i1}, \bar{{\bf{R}}}^{l}_{i2}, \cdots,\bar{{\bf{R}}}^{l}_{i j}]
\end{align}
where $[,]$ denotes the concatenate operation.
The self-representation results are generated by stacking all feature maps, which is given by:
\begin{align}
  \bar{{\bf{U}}}^{l} = [\bar{{\bf{U}}}^{l}_{1}; \bar{{\bf{U}}}^{l}_{2};\cdots;\bar{{\bf{U}}}^{l}_{s}] \\
  \bar{{\bf{R}}}^{l} = [\bar{{\bf{R}}}^{l}_{1}; \bar{{\bf{R}}}^{l}_{2};\cdots;\bar{{\bf{R}}}^{l}_{s}]
\end{align}
where $s$ is the number of the filters and $[;]$ represents the stack operation.

\subsubsection{Cross-Representation}
We use a scaled dot-product attention \citep{vaswani2017attention} as the implementation of the cross-attention module.
Formally, given queries $\bf{Q}$, keys $\bf{K}$ and values $\bf{V}$ as three matrices with multiple stacked word vectors, the dot-product attention outputs the weighted sums $\hat{\bf{V}}$ by:
\begin{align}
  \hat{\bf{V}} = softmax(\frac{{\bf{Q}} {\bf{K}}^{T}} {\sqrt{d}}) \cdot {\bf{V}}
\end{align}
For convenience, we denote the scaled dot-product attention as $f_{att}$.
The cross-representation results are defined as:
\begin{align}
  \hat{{\bf{U}}}^{l} = f_{att} (F_{1}({\bf{U}}^{l}), F_{2}({\bf{R}}^{l}), {\bf{R}}^{l}) \\
  \hat{{\bf{R}}}^{l} = f_{att} (F_{2}({\bf{R}}^{l}), F_{1}({\bf{U}}^{l}), {\bf{U}}^{l})
\end{align}
where $F_{1}$, $F_{2}$ are single-layer feed-forward networks which map ${\bf{U}}^{l}$ and ${\bf{R}}^{l}$ into a same latent space.

\subsubsection{Fusion}
Having the two types of representations at hand, the fusion module fuses the representations and the input embeddings in two steps.
Take $\bar{\bf{U}}^{l}, \hat{\bf{U}}^{l}$ as an example.
The fusion module firstly fuses each representation with the corresponding input embedding following \citep{mou2015natural}:
\begin{align}
  \tilde{{\bf{U}}}^{l}_{1} = G_{1} ([{\bf{U}}^{l}, \bar{{\bf{U}}}^{l} , {\bf{U}}^{l} - \bar{{\bf{U}}}^{l}, {\bf{U}}^{l} \odot \bar{{\bf{U}}}^{l}]) \\
  \tilde{{\bf{U}}}^{l}_{2} = G_{2} ([{\bf{U}}^{l}, \hat{{\bf{U}}}^{l} , {\bf{U}}^{l} - \hat{{\bf{U}}}^{l}, {\bf{U}}^{l} \odot \hat{{\bf{U}}}^{l}])
\end{align}
where $\odot$ denotes the element-wise multiplication.
Then the results are further fused by:
\begin{align}
  \tilde{{\bf{U}}}^{l} = G ([\tilde{{\bf{U}}}^{l}_{1} , \tilde{{\bf{U}}}^{l}_{2}] )
\end{align}
where $G_1, G_2, G$ are single-layer feed-forward networks with independent parameters.
The operations for $\bar{{\bf{R}}}^{l}$ and $\hat{{\bf{R}}}^{l}$ are the same and here we omit the details.
The fusion results are then outputted as two new representations for the two input sentences.

We add a residual connection \citep{he2016deep} and a direct connection \cite{Tao:2019b} from the word embedding layer to each block, whose outputs together with $\tilde{\bf{U}}^{l}$ and $\tilde{\bf{R}}^{l}$ will then be performed the layer normalization \citep{ba2016layer} and transformed to ${\bf{U}}^{l+1}$ and ${\bf{R}}^{l+1}$.

\subsection{Matching}
\label{section43}
As it is shown in Figure \ref{figure:sentence_block}, S2M directly generates the sentence vectors ${\bf{v}}^{l}_{u}$ and ${\bf{v}}^{l}_{r}$ of utterance and response respectively through a pooling operation on $\tilde{{\bf{U}}}^{l}$ and $\tilde{{\bf{R}}}^{l}$.
Then an matching oriented interaction operation \citep{mou2015natural, Yang:2019} is performed on the sentence vectors to obtain the matching feature:
\begin{align}
  {\bf{m}}^{l} = H([{\bf{v}}^{l}_{u}, {\bf{v}}^{l}_{r}, {\bf{v}}^{l}_{u} - {\bf{v}}^{l}_{r}, {\bf{v}}^{l}_{u} \odot {\bf{v}}^{l}_{r}] )
\end{align}
$H$ is a single-layer feed-forward network.
We conduct further analysis for different pooling strategies, as will be discussed in the following section.

\subsection{Aggregation}
After the matching stage, we have collected a list of multi-turn matching features.
The features in the same stack are aggregated and fed into a GRU \citep{chung2014empirical} to capture the temporal relationship.
Let ${\bf{h}}^{l}_{i}$ and ${\bf{m}}^{l}_{i}$ be the $i$-th hidden state and matching feature in the $l$-th stack, ${\bf{h}}^{l}_{i}$ is defined as:
\begin{align}
  {\bf{h}}^{l}_{i} = \text{GRU}({\bf{h}}^{l}_{i-1}, {\bf{m}}^{l}_{i})
\end{align}

The matching score for the $l$-th stack is calculated by:
\begin{align}
g^{l}(c, r) = \sigma ({\bf{W}} \cdot {\bf{h}}^{l}_{final} + {\bf{b}})
\end{align}
where ${\bf{h}}^{l}_{final}$ is the final output of GRU in the $l$-th stack, ${\bf{W}}$ and $\bf{b}$ are trainable parameters and $\sigma$ represents the sigmoid function.
Following \citep{Tao:2019b}, we simply add $g^{l}(c, r)$ of each stack to form the final matching score:
\begin{align}
  g(c, r) = \sum_{l=1}^{L} g^{l}(c, r)
\end{align}
where $L$ represents the stack number and is tuned manually as a hyper-parameter.

\begin{figure}[t]
    \centering
    \includegraphics[width=0.45\textwidth]{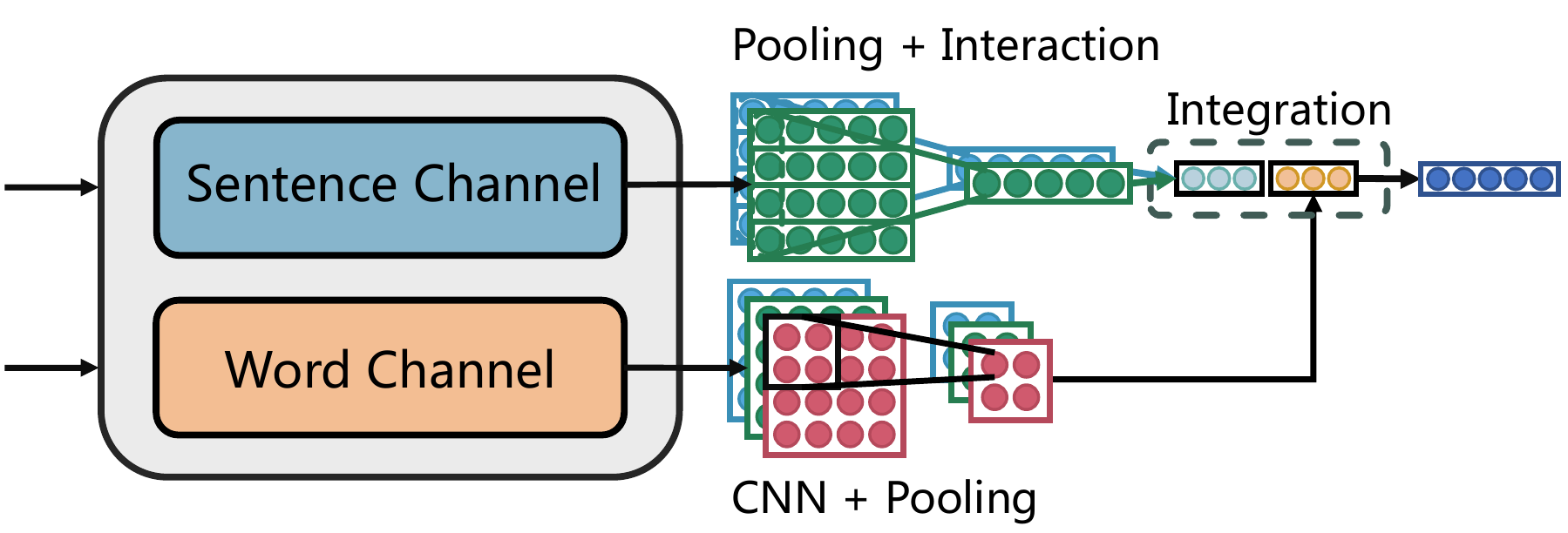}
    \caption{Integration of matching at sentence and word-level.}
    \label{figure:fusion}
\end{figure}

\subsection{Integration}
\label{section45}
We further integrate the sentence-level matching with the word-level matching \citep{Zhou:2018, Tao:2019b} as shown in Figure \ref{figure:fusion}.
We call the two lines of operations in the figure two ``channels'' for narrative convenience.
The word matching channel first generates word similarity matrices $\bf{M}$ based on self-attention and cross-attention representations, and then applies CNN \citep{krizhevsky2012imagenet} and pooling layer to extract the matching features.
Details can be found in the two referenced studies.
The sentence matching channel is as described in Section \ref{section42} and \ref{section43}.

We explore three different integration strategies to see how the two channels benefit from each other.
In the first strategy $I_1$, we integrate them at the representation stage.
The original inputs for each channel are directly concatenated to get new inputs, which are fed to each channel instead.
In the second strategy $I_2$, the integration happens at the matching stage, which is to say, 
the results of the self-representation and cross-representation modules in the sentence-level matching channel are multiplied to form word similarity matrices, 
while the results of the fusion module in the word-level matching channel are pooled to form two sentence vectors.
Then we stack the new word similarity matrices on $\bf{M}$, and concatenate the new sentence vectors to ${\bf{v}}^{l}_{u}$ and ${\bf{v}}^{l}_{r}$ respectively.
For the first two strategies, we use independent GRU and loss for each channel, and sum up the losses as the total loss.
The third strategy $I_3$ integrates the matching information at the aggregation stage.
The matching features from both channels are concatenated before they are fed into the shared GRU.
We compare these strategies through empirical studies. The results will be reported in the Section \ref{section:experiment}.

\subsection{Learning}
We typically use the negative log likelihood as the loss function in each stack.
Following \citep{Tao:2019b}, we sum up all losses in each stack to form the total loss:
\begin{align}
\begin{split}
  -\sum_{l=1}^{L} \sum_{k=1}^{K} & [y_{k} \log(g^{l}(c_{k}, r_{k})) \\
          &+ (1 - y_{k}) \log(1 - g^{l}(c_{k}, r_{k}))]
\end{split}
\end{align}

\begin{table*}
\centering
\resizebox{0.90\textwidth}{!}
  {\begin{tabular}{l|c c c|c c c c c c|c c c}
    \toprule
    \multirow{2}{*}{Models} & \multicolumn{3}{c|}{Ubuntu Corpus} & \multicolumn{6}{c|}{Douban Corpus} & \multicolumn{3}{c}{Ecommerce Corpus}\\
    %\midrule{2-13}
      & $\textbf{R}_{10}@1$ & $\textbf{R}_{10}@2$ & $\textbf{R}_{10}@5$ & \textbf{MAP} & \textbf{MRR} & \textbf{P}@1 & $\textbf{R}_{10}@1$ & $\textbf{R}_{10}@2$ & $\textbf{R}_{10}@5$ & $\textbf{R}_{10}@1$ & $\textbf{R}_{10}@2$ & $\textbf{R}_{10}@5$\\
    \midrule
    TF-IDF  & 0.410 & 0.545 & 0.708 & 0.331 & 0.359 & 0.180 & 0.096 & 0.172 & 0.405 & 0.159 & 0.256 & 0.477\\
    CNN     & 0.549 & 0.684 & 0.896 & 0.417 & 0.440 & 0.226 & 0.121 & 0.252 & 0.647 & 0.328 & 0.515 & 0.792\\
    BiLSTM  & 0.630 & 0.780 & 0.944 & 0.479 & 0.514 & 0.313 & 0.184 & 0.330 & 0.716 & 0.365 & 0.536 & 0.828\\
    \midrule
    Multi-View & 0.662 & 0.801 & 0.951 & 0.505 & 0.543 & 0.342 & 0.202 & 0.350 & 0.729 & 0.421 & 0.601 & 0.861\\ 
    DL2R       & 0.626 & 0.783 & 0.944 & 0.488 & 0.527 & 0.330 & 0.193 & 0.342 & 0.705 & 0.399 & 0.571 & 0.842\\
%    IMN        & 0.777 & 0.880 & 0.974 & 0.570 & 0.615 & 0.433 & 0.262 & 0.452 & 0.789 & 0.621 & 0.797 & 0.964\\
    \midrule
    SMN  & 0.726 & 0.847 & 0.961 & 0.529 & 0.569 & 0.397 & 0.233 & 0.396 & 0.724 & 0.453 & 0.654 & 0.886\\
    DUA  & 0.752 & 0.868 & 0.962 & 0.551 & 0.599 & 0.421 & 0.243 & 0.421 & 0.780 & 0.501 & 0.700 & 0.921\\
    DAM  & 0.767 & 0.874 & 0.969 & 0.550 & 0.601 & 0.427 & 0.254 & 0.410 & 0.757 & 0.526 & 0.727 & 0.933\\
    IOI  & 0.796 & 0.894 & 0.974 & 0.573 & 0.621 & 0.444 & 0.269 & 0.451 & 0.786 & 0.563 & 0.768 & 0.950\\
    \midrule
    $\text{S2M}_{pure}$ & \bf{0.807} & \bf{0.898} & \bf{0.976} & \bf{0.574} & 0.620 & \textbf{0.451} & \bf{0.273} & 0.451 & \bf{0.787} & \bf{0.598} & \bf{0.777} & \bf{0.950}\\
    $\text{S2M}_{I_1}$ & 0.793 & 0.892 & 0.973 & 0.563  & 0.607 & 0.434 & 0.261 & 0.444 & 0.771 & 0.568 & 0.738 & 0.941\\
    $\text{S2M}_{I_2}$ & 0.808 & 0.900 & 0.977 & 0.574  & 0.619 & 0.452 & 0.276 & 0.451 & 0.786 & 0.601 & 0.775 & 0.949\\
    $\text{S2M}_{I_3}$ & \bf{0.813} & \bf{0.903} & \bf{0.978} & \bf{0.579}  & \bf{0.630} & \bf{0.466} & \bf{0.283} & \bf{0.456} & \bf{0.792} & \bf{0.619} & \bf{0.807} & \bf{0.958}\\
    \bottomrule
  \end{tabular}}
  \caption{Experiment results of our model and other baseline methods on three benchmark datasets. The subscript $pure$ represents the model only containing sentence matching; $I_1$, $I_2$ and $I_3$ represents the model integrates with IOI in representation, matching and aggregation stages respectively.}
  \label{table:overall_result}
\end{table*}

\section{Experiment}
\label{section:experiment}
\subsection{Datasets}
We test our model on three benchmark datasets: Ubuntu Dialogue Corpus \cite{Lowe:2015}, Douban Conversation Corpus \cite{Wu:2017} and E-commerce Dialogue Corpus \cite{Zhang:2018b}.
Ubuntu Dialogue Corpus contains multi-turn dialogues on technical support collected from chat logs of the Ubuntu forum.
The corpus consists of 1 million context-response pairs for training and 0.5 million for validation and testing and the ratios of positive samples and negative samples in three sets are 1:1, 1:9 and 1:9 respectively. 
All the positive samples are corresponding responses in the dialogues while the negative samples are randomly chosen from the corpus.

Douban Conversation Corpus is a Chinese multi-turn conversations dataset, which is crawled from a Chinese social network on open domain topics \footnote{https://www.douban.com/group}.
There are 1 million instances for training, 50 thousand instances for validation and 6670 instances for testing.
In training and validation set, the last utterance in each context is used as a positive response and a random utterance of the corpus is sampled as a negative response.
However, in the test set, the candidate responses are all retrieved via an inverted-index system (Lucene\footnote{https://lucenenet.apache.org/}).
The label of these candidates are annotated by human. 
Therefore, a context of the test set in Douban Conversation Corpus may have more than one positive responses.

E-commerce Dialogue Corpus is another Chinese dataset consisting of multi-turn real-world conversations between customers and customer service representatives from Taobao\footnote{https://www.taobao.com}.
It contains 1 million pairs for training and 10 thousand pairs for validation and testing.
The ratio of positive and negative pairs is 1:1 for training and validation, and 1:9 for testing.
The negative responses are constructed by ranking the response pool based on the last utterance along with top-5 key words in the context.

\subsection{Metrics}
Following the previous work \cite{Wu:2017}, we calculate the recall of true positive responses among the top-$k$ selected responses from $n$ available candidates for the given conversation, denoted as $\textbf{R}_n@k$.
Since there exists more than one positive response in Douban Corpus and $\textbf{R}_{n}@k$ is not sufficient for evaluation, we take mean average precision (\textbf{MAP}) and mean reciprocal rank (\textbf{MRR}) into consideration, which show the potential of a system providing more than one candidate response \cite{Yan:2016}.
Moreover, we use a natural evaluation metric, precision at the 1st position (\textbf{P}@1), as another metric for references.

\subsection{Baseline Methods}
In this paper, the baseline methods for comparison are as follows:

\textbf{Basic Models}: Basic models in \cite{Lowe:2015,Kadlec:2015} including TF-IDF, CNN and BiLSTM in early works.

\textbf{Global Matching Models}: DL2R\cite{Yan:2016} and Multi-view\cite{Zhou:2016}.
These models represent context and response as two vectors and use them to calculate the final matching score.
In this paper, we does not compare with IMN\cite{Gu:2019}, because it introduces some word embeddings including the 300-dimensional GloVe embeddings \cite{pennington2014glove} and the 200-dimensional embeddings \cite{song2018directional} pretrained on external corpora, which is unfair to compare with our model and others using Word2Vec only.

\textbf{Sequential Matching Models}: SMN\cite{Wu:2017}, DUA\cite{Zhang:2018b}, DAM\cite{Zhou:2018} and IOI\cite{Tao:2019b}.
These methods construct similarity matrix between response and each utterance in context based on the representation.
Then, a CNN + RNN or a 3D-CNN is applied to calculate the final matching score.

\subsection{Implementation Details}
We implement our models in Tensorflow \cite{abadi2016tensorflow}.
In data preprocessing, we limit the maximum utterance or response length to 50 words.
The maximum context length is set to 15.
Truncating or zero-padding is applied to a context or response if necessary.
In model construction, we set the stack number as 7.
The word embeddings are pre-trained with Word2Vec algorithm \cite{mikolov2013distributed} on corresponding training set of corpus and the word vector dimension is set to 200.
In self-representation, we use a one-layer 1D convolution network and the kernel size is set to 3 with stride as 1.
The hidden size and the dimension of hidden states in GRU are both set to 200.
For the integration of the three strategies, we choose IOI\cite{Tao:2019b} to serve as the word channel, which is a state-of-the-art model under the sequential matching framework of SMN.
In optimization, Adam \cite{kingma2014adam} is introduced to update all of parameters.
The batch size is set to 20.
The learning rate is initialized as 5$\times\text{10}^{\text{-4}}$ and exponentially decayed every 5000 steps.
We freeze the word embedding during training.
Our model achieves the best result when traversing approximately 2 epochs of the whole training samples in Ubuntu Dialogue Corpus and 4 epochs in Douban Conversation Corpus and Ecommerce Dialogue Corpus.

%\vspace{-0.5em}
\subsection{Evaluation Results}
Table \ref{table:overall_result} displays the results of our proposed model S2M and all other baselines. 
All the results except ours are directly from the corresponding papers.
From the table, we can see that $\text{S2M}_{pure}$ achieves a competitive result on three datasets.
This result is a strong evidence showing that sequential sentence matching is another effective method in multi-turn response selection task.
Moreover, $\text{S2M}_{I_3}$ is consistently better than the current best performing method by a margin of 1.7\% in terms of $\textbf{R}_{10}@1$ on Ubuntu Dialogue Corpus; 0.9\% in terms of \textbf{MRR} and 1.4\% in terms of $\textbf{R}_{10}@1$ on Douban Conversation Corpus; and 5.6\% in terms of $\textbf{R}_{10}@1$ on E-commerce Dialogue Corpus, achieving a new state-of-the-art performance on three datasets.
It should be noticed that the negative candidates in E-commerce Dialogue Corpus are retrieved according to some key words in the context, which strongly confuses the models matching in word-level and segment-level with similarity matrices. Such a great margin of improvement can strongly support our standpoint that sentence matching mechanism can address the problem that we discussed in the introduction.

Besides, when making a deep insight into the integration strategies, we can find $\text{S2M}_{I_1} < \text{S2M}_{pure} < \text{S2M}_{I_2} < \text{S2M}_{I_3}$ from the table.
As we know, word or segment representation and sentence-level representation are in different semantic spaces.
We believe integration in the representing stage will break their independence, thus their matching procedures will be weaken and become vague, leading to a performance drop.
However, if we keep the representations independently and do the integration in matching or aggregation stage, the results have an improvement and the improvement is maximized in aggregation stage.
We think in matching stage, each channel keeps the matching information of their own and have some supplementary information from another channel, thus the result has a slight improvement.
When integrating in aggregation stage, both of the features are expressions of matching information over the utterance and response from two separate aspects and they can share the temporal relationships along the conversation, leading to an effective improvement.

\vspace{-0.5em}
\subsection{Further Analysis}
We use the Ubuntu Dialogue Corpus to carry out some detailed experiments to analyze S2M.

\begin{table}
\resizebox{0.45\textwidth}{!}
{
\begin{tabular}[!tbp]{c l r r r}
\toprule
& Models & $\textbf{R}_{10}@1$ & $\textbf{R}_{10}@2$ & $\textbf{R}_{10}@5$ \\
\midrule
& $\text{S2M}_{pure}$ &  0.807 &   0.898  & 0.976  \\
\midrule
\multirow{2}{*}{1st} & $\text{S2M}_{cross}$ &  0.799 &  0.896 & 0.977  \\
& $\text{S2M}_{self}$ &  0.791 &  0.890  & 0.975  \\
\midrule
\multirow{2}{*}{2nd} & $\text{S2M}_{GRU\ rep}$ &  0.807  &  0.900  & 0.977  \\
& $\text{S2M}_{Att\ rep}$ &  0.800  & 0.898  & 0.977  \\
\midrule
\multirow{2}{*}{3rd} &  $\text{S2M}_{Mean\ sen}$ &  0.801  &  0.896  & 0.976  \\
& $\text{S2M}_{GRU\ sen}$ &  0.805  & 0.897  & 0.977  \\
\bottomrule
\end{tabular}}
\caption{Ablation Experiment Results. $\text{S2M}_{cross}$ and $\text{S2M}_{self}$ in the first part mean that we only use cross/self representation component. In the second part, GRU or attention mechanism replaces the 1D-CNN to get the self representation, denoted as $\text{S2M}_{GRU\ rep}$ and $\text{S2M}_{att\ rep}$. $\text{S2M}_{GRU\ sen}$ and $\text{S2M}_{Mean\ sen}$ in the third part indicates using GRU or MeanPooling method to get the sentence representation.} 
\label{table-ablation}
\end{table}

\textbf{Model Ablation}: 
We explore the effect of each component in our model by removing or replacing them one by one from $\text{S2M}_{pure}$.
Firstly, we remove the self-representation and cross-representation components respectively.
The first part in Table \ref{table-ablation} indicates both representation components play important roles in S2M.
Cross-representation contributes a little bit more than self-representation, which states the necessity of jointly considering the relationship of utterance and response.
We also carry out experiments to study two different kinds of self-representation implementations including GRU and Attention mechanism.
The second part in Table \ref{table-ablation} implies CNN $\approx$ GRU $>$ Attention mechanism.
In our opinion, Attention mechanism without position embedding is a kind of ``bag-of-word'' model \cite{vaswani2017attention} and it lacks the order information.
Meanwhile both CNN and GRU are natural methods to make use of the order relationships in sentence.
Finally, we extend the method of obtaining the sentence representation in matching component.
In the third part of Table \ref{table-ablation}, we can see that GRU and Max Pooling lead to similar results and are both better than Mean Pooling.
This is also reasonable since GRU has independent parameters and the max operation keeps the maximal value along the sentence dimension, and thus they both can maintain the core part of each utterance and candidate response.

\begin{figure}[t]
    \centering
    \subfigure[]{\includegraphics[width=0.36\textwidth]{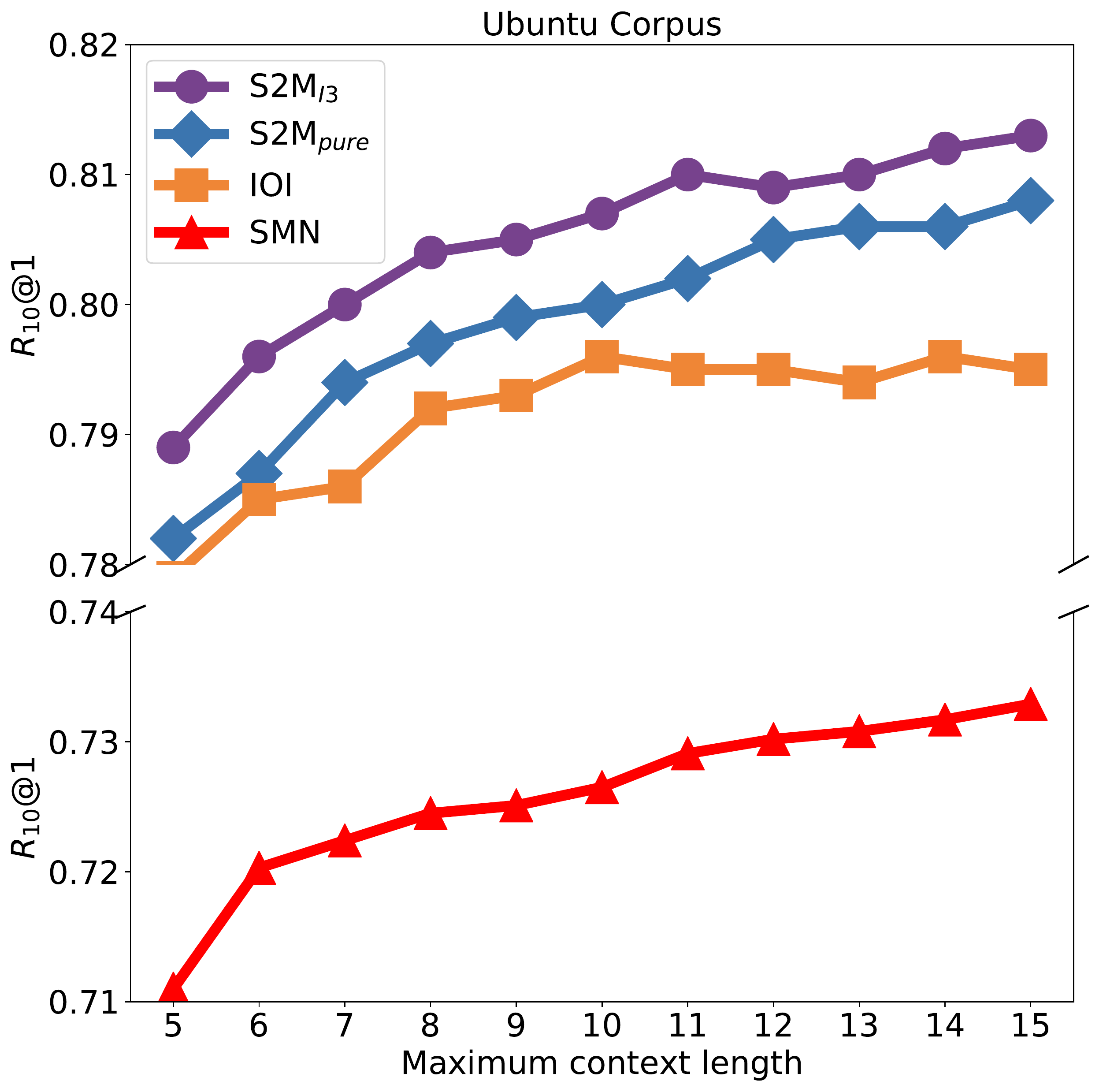}\label{figure:a}}
    \subfigure[]{\includegraphics[width=0.36\textwidth]{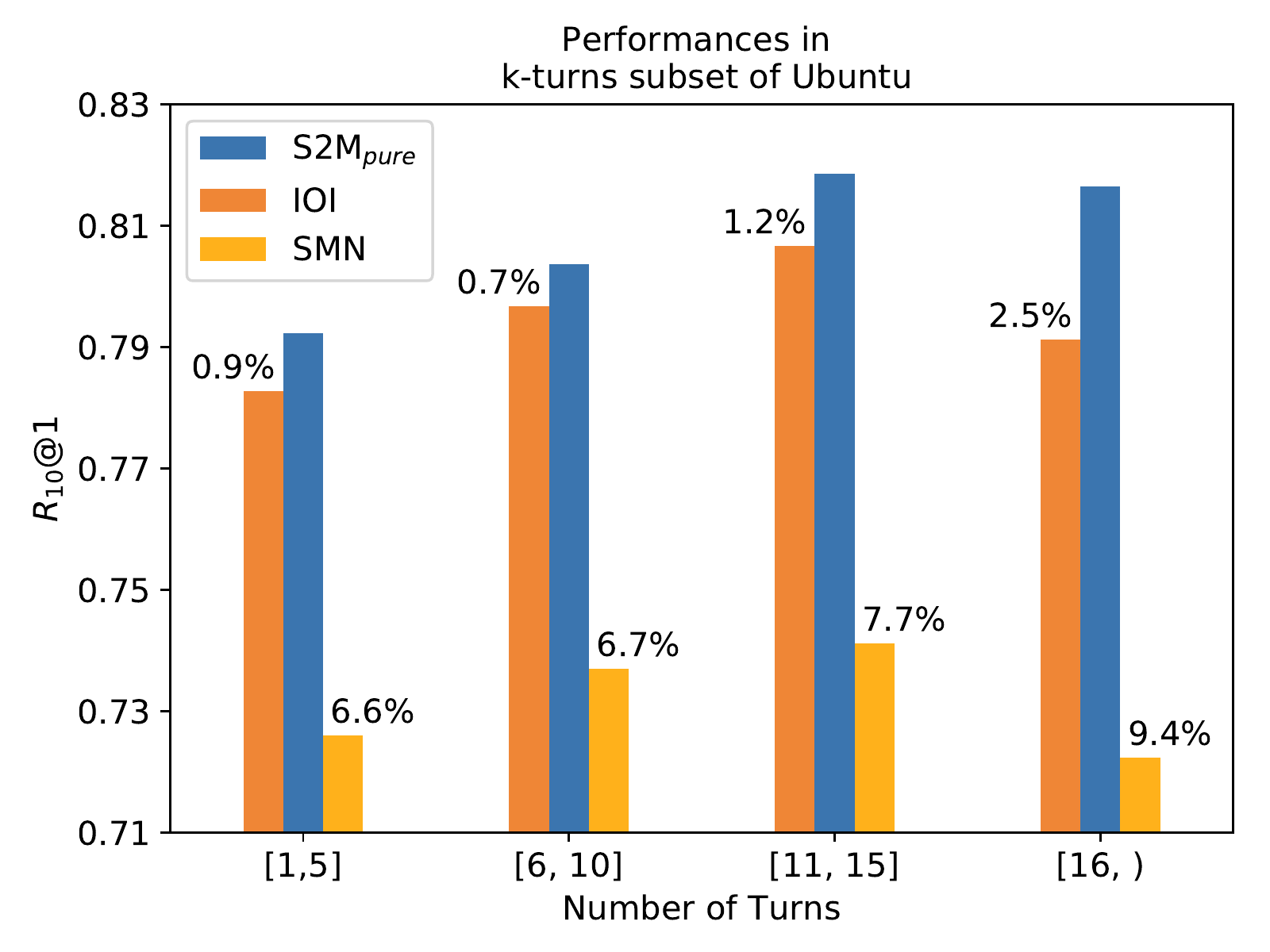}\label{figure:b}}
    \caption{Performance of models on Ubuntu Dialogue Corpus. \ref{figure:a} shows the result of S2M with different numbers of turns. \ref{figure:b} gives a detailed comparisons on different numbers of turns in the test set.}
    \label{figure:turn-numbers}
\end{figure}
\textbf{Quantity Analysis}:
We further study how the number of turns of dialogues affects our model.
To provide a more persuasive result, we compare S2M with SMN\cite{Wu:2017} and IOI\cite{Tao:2019b} which are the representative and state-of-the-art models based on word-level and segment-level matching. 
As the results reported in their papers are not enough for use, we run their open source codes \footnote{https://github.com/chongyangtao/IOI} \footnote{https://github.com/MarkWuNLP/MultiTurnResponseSelection}with the recommended hyperparameters on various turns.
Constrained by the machine resources, we finally set the maximum context length as 15.
Figure \ref{figure:a} presents the comparison of performance with different numbers of turns of dialogues.
From the chart, we have some observations: 
(1) as the number of turns increases, the performance improves for all of the models;
(2) the performance of IOI increases more gently as the number of turns approaches 11 and its best result is achieved at 10 turns;
(3) $\text{S2M}_{pure}$ and $\text{S2M}_{I_3}$ can still benefit from increasing the number of turns up to 15 even though with a high performance (more than 0.81 of $\textbf{R}_{10}@1$).
The result implies our model is statistically more competent to deal with longer conversations. Moreover, we design another experiment to find out where the improvement comes from. 
We divide the testing set of Ubuntu Dialogue Corpus into several subsets according to the number of turns and report the performances of SMN, IOI and S2M in Figure \ref{figure:b}.
In the figure, we can find that the gap of the performance becomes larger in the subsets when the context length is increasing.

We think our model performs better on longer context because it benefits from the sentence-level information.
It concentrates on the sentence information extraction and can neglect similar words in both utterance and response.
Such words are just a kind of repetition over the utterance and response, which may not constitute a reply relationship and would have a relatively large negative effect if matching is carried out based on the word-level similarity matrix, as SMN.
Thus, our models can make a full use of the extra information brought by the context without suffering from unexpected word matching, so that it does better in long-turn response selection.

\vspace{-0.5em}
\section{Conclusion and Future Work}
In this paper, we propose a sequential sentence matching network (S2M), which reduces the word-level matching errors in multi-turn response selection using the sentence-level semantic information.
We also find that integration of sentence-level matching with word-level matching enables our model to learn richer features.
Evaluation results on benchmark datasets indicate that our model can significantly outperform the state-of-the-art models.
Moreover, for longer dialogues, our model is more competent than the state-of-the-art models that are based on word-level or segment-level matching.
This further demonstrates the advantage of using sentence-level semantic information on the multi-turn response selection task.
In the future, we will look into the remaining bad cases of our model of different corpora and try to identify further improvement opportunities.

\bibliography{acl2020}
\bibliographystyle{acl_natbib}

\end{document}